\documentclass[10pt,twocolumn,letterpaper]{article}

\usepackage[pagenumbers]{cvpr} % To force page numbers, e.g. for an arXiv version
\usepackage{appendix}

\usepackage[dvipsnames]{xcolor}

\definecolor{cvprblue}{rgb}{0.21,0.49,0.74}
\usepackage[pagebackref,breaklinks,colorlinks,citecolor=cvprblue]{hyperref}
\usepackage[symbol]{footmisc}

\def\paperID{181} % *** Enter the Paper ID here
\def\confName{3DV\xspace}
\def\confYear{2024\xspace}

\newcommand{\velocity}{\Delta}
\newcommand{\scan}{\mathcal{S}}
\newcommand{\joints}{\mathcal{J}}
\newcommand{\skeletonextract}{\text{skeleton}}
\newcommand{\retargetrnn}{\text{SMRM}}
\newcommand{\rotation}{\theta}
\newcommand{\blpose}{o}
\newcommand{\blposenet}{O_{net}}
\newcommand{\weightsnet}{W_{net}}
\newcommand{\weights}{W}
\newcommand{\motionloss}{\mathcal{L}_{motion}}
\newcommand{\geometryloss}{\mathcal{L}_{geom}}

\usepackage{pifont}

\title{Correspondence-Free Online Human Motion Retargeting}

\author{Rim Rekik\textsuperscript{*}\textsuperscript{1}~~~%\\
Mathieu Marsot\textsuperscript{*}\textsuperscript{1}~~~%\\
Anne-H\'{e}l\`{e}ne Olivier\textsuperscript{2}~~~%\\
Jean-S\'{e}bastien Franco\textsuperscript{1}~~~%\\
Stefanie Wuhrer\textsuperscript{1}\\
\textsuperscript{1} \small{Univ. Grenoble Alpes, Inria, CNRS, Grenoble INP \textsuperscript{†}, LJK, 38000 Grenoble, France}\\
\textsuperscript{2} \small{Univ. Rennes, Inria, CNRS, IRISA, M2S, 35000 Rennes, France}\\
\textsuperscript{*} \tt\small Joint first authorship: rim.rekik-dit-nekhili@inria.fr, mathieu.marsot@gmail.com\\
{\tt\small firstname.lastname@inria.fr}
}

\begin{document}

\maketitle
\footnotetext[2]{Institute of Engineering Univ. Grenoble Alpes
\label{footnote}}

\begin{abstract}
We present a data-driven framework for unsupervised human motion retargeting that animates a target subject with the motion of a source subject. Our method is correspondence-free, requiring neither spatial correspondences between the source and target shapes nor temporal correspondences between different frames of the source motion. This allows to animate a target shape with arbitrary sequences of humans in motion, possibly captured using 4D acquisition platforms or consumer devices. Our method unifies the advantages of two existing lines of work, namely skeletal motion retargeting, which leverages long-term temporal context, and surface-based retargeting, which preserves surface details, by combining a geometry-aware deformation model with a skeleton-aware motion transfer approach. This allows to take into account long-term temporal context while accounting for surface details.
During inference, our method runs online,~\ie input can be processed in a serial way, and retargeting is performed in a single forward pass per frame.
Experiments show that including long-term temporal context during training improves the method's accuracy for skeletal motion and detail preservation. Furthermore, our method generalizes to unobserved motions and body shapes. We demonstrate that our method achieves state-of-the-art results on two test datasets and that it can be used to animate human models with the output of a multi-view acquisition platform. Code is available at \url{https://gitlab.inria.fr/rrekikdi/human-motion-retargeting2023}.

\begin{figure}
    \centering
    \includegraphics[width=\columnwidth]{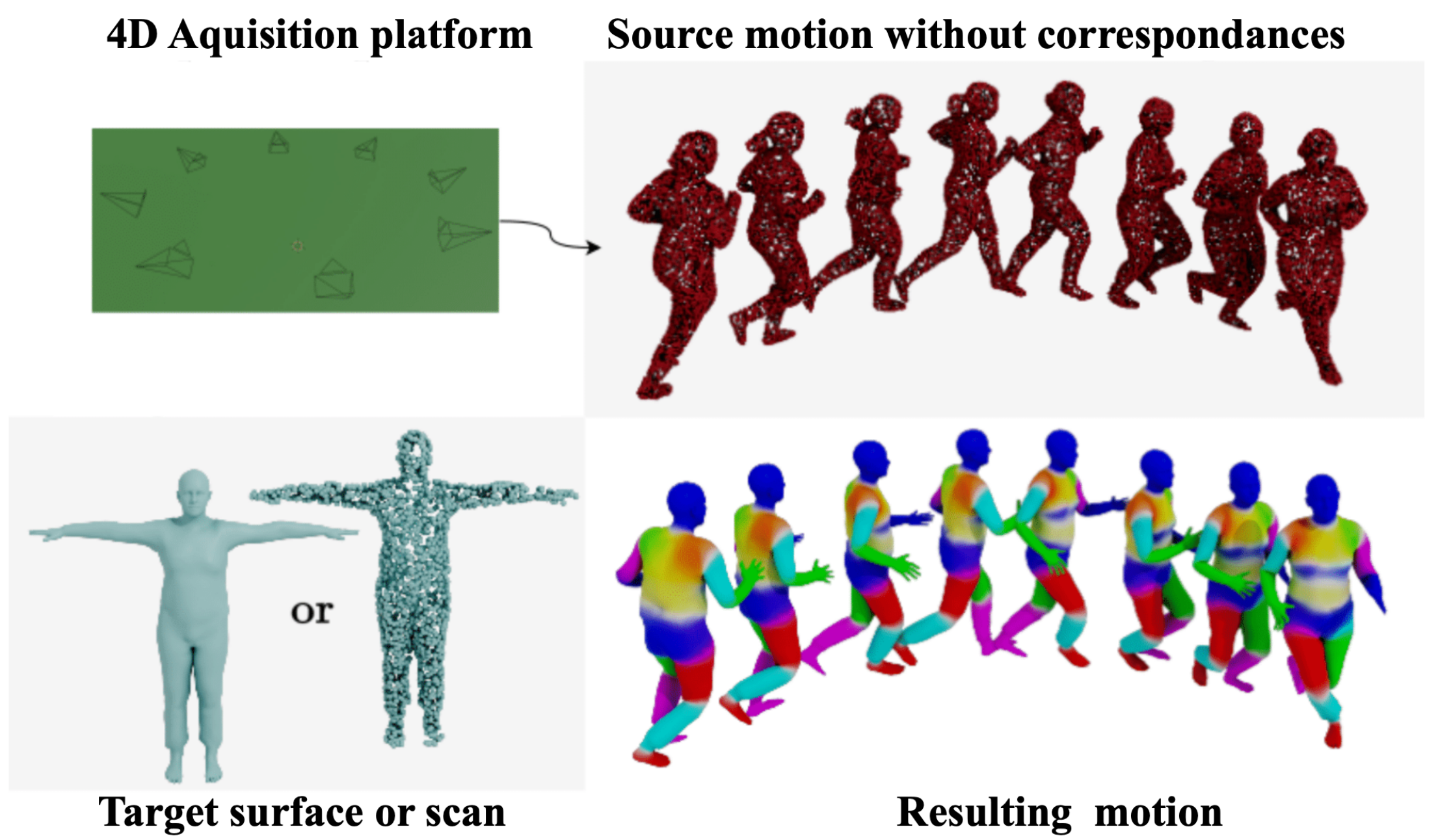}
    \caption{Given an untracked motion performed by a source subject (top) and a target body shape (bottom left), our method animates the target with the source motion, preserving temporal correspondences of the output motion (bottom right).%\ah{ça serait attractif d'avoir une vraie image de teaser qui prend toute la largeur de la page}
    }
    \label{fig:teaser}
\end{figure}

\end{abstract}

\section{Introduction}
\label{sec:intro}

Human motion retargeting is the process of animating a target character with the motion of a source character. We study this problem for densely sampled 3D surfaces, and arbitrary motion duration. This has applications in video gaming, movie making, avatar animation, and augmenting existing datasets of 4D body motions~\eg~\cite{bogo2014faust, bogo2017dynamic, von2018recovering, mahmood2019amass}. 

Motion retargeting has recently made significant progress. Skeleton-based methods can take long-term temporal context of about $2$ seconds into account~\eg~\cite{gleicher1998retargetting, villegas2018neural, lim2019pmnet}. Surface deformation-based methods allow to retarget geometric details while possibly capturing short time dynamics~\eg~\cite{sumner2004deformation, wang2020neural, basset2021neural, regateiro2022temporal}.

Most of these works take as input structured data such as skeletons or template-aligned surfaces for which correspondences are known. 

The goal of this work is correspondence-free retargeting that takes as input unstructured 4D data for which neither correspondences between the source and target shape nor correspondences between different frames of the source motion are given. This allows to directly animate a human target shape with arbitrary sequences of humans in motion, possibly captured using 4D acquisition platforms. Solving the retargeting problem for unstructured 3D data is challenging as computing correspondences is a combinatorial problem. Fig.~\ref{fig:teaser} illustrates the problem's input and output. 

Recently, a first solution for correspondence-free motion retargeting has been proposed~\cite{jiang2022h4d}, which introduces a generic motion prior to describe how body shapes move in a low dimensional space, and takes long-term temporal context into account. While this opens the door to novel solutions, during inference the method is limited to retarget sequences of fixed length. Inspired by this work, our framework also learns long-term temporal context.

In this work, we propose the first correspondence-free method for online motion retargeting that learns long-term temporal context. Our method is able to learn temporal context of $1s$ duration, which we demonstrate to improve the accuracy of motion retargeting on challenging examples.

We achieve this by combining the advantages of skeletal and surface-based motion retargeting to learn both temporal context and geometric detail. More specifically, we combine skeleton-based motion retargeting and shape deformation modeling, which allows to handle high-dimensional spatio-temporal data. The reason is that skeletal retargeting methods achieve good results by encoding long-term temporal context~\cite{villegas2018neural} and that recent methods to deform the shape based on skeletal deformations ~\cite{liu2019neuroskinning,xu2020rignet,yang2021s3,mosella2022skinningnet} can accurately model geometric detail. 
Our hypothesis is that the locomotion information of the source sequence is explained at the skeletal level, while the dense surface details are intrinsic to the target shape. Thus, our idea is to minimize a motion preservation loss that transfers the skeletal motion of the source to the target, and a shape preservation loss that keeps the shape details of the target.

Our main contributions are:
\begin{itemize}
\item The first correspondence-free online approach for dense human motion retargeting that learns temporal context.
\item A demonstration that long-term temporal context improves the accuracy of motion retargeting.
\item An approach that outperforms correspondence-free methods in both geometric detail preservation and skeletal-level motion retargeting. 

\end{itemize}

\section{Related Work}

Works addressing motion retargeting can be categorized into three groups: works operating at the skeleton level, works transferring shape deformations at a dense geometric level, and motion priors applied to motion retargeting. 

\subsection{Skeletal motion transfer}
Early works for motion retargeting are based on skeleton parametrizations, where human pose is described with a sparse set of joint locations. Gleicher~\etal~\cite{gleicher1998retargetting} introduced this task, considering it as an optimization problem  with kinematic constraints over the entire motion sequence. The resulting skeletal parametrizations can then be animated with manually computed skinning weights~\cite{lipman2005linear,yu2004mesh} and using blending techniques~\cite{joshi2006learning,kry2002eigenskin}.

Follow up works~\cite{lee1999hierarchical,choi2000online,tak2005physically,ko2003research} also require iterative optimization with hand-designed kinematic constraints for particular motions.
With the surge accessibility of captured motion data and the efficiency of deep learning techniques, new data-driven approaches~\cite{delhaisse2017transfer,jang2018variational,villegas2018neural,aberman2020skeleton} have shown outstanding results without requiring many handcrafted energies. 
Some of these data-driven approaches~\cite{delhaisse2017transfer,jang2018variational} require paired training data, whose design involves human effort. Therefore, another line of works~\cite{villegas2018neural,aberman2020skeleton} propose unsupervised retargeting strategies. Villegas~\etal~\cite{villegas2018neural} propose an unsupervised motion retargeting framework based on adversarial cycle consistency to ensure plausibility of the retargeted motions. This method generates natural motions for unseen characters, but only operates at a skeletal level and does not include geometric details.

More recent work~\cite{villegas21} proposes to include geometry and investigates hybrid skeleton-based motion retargeting with both a data-driven network and a post inference optimization to preserve self-contacts and prevent interpenetration. This method shows outstanding results, yet requires manually handcrafted skinning weights as additional input. 

Similar to our approach, a recent method combines skeletal-level and geometric information for motion retargeting~\cite{zhang2023skinned}. Unlike our work, this method focuses on contact preservation, in particular for characters created with computer aided design tools for which clean skeletal models are provided as input. In contrast, we focus on retargeting the raw output from acquisition platforms directly, without the need for costly and error-prone pre-processing to find correspondence information or fit skeletons.

\subsection{Shape deformation transfer}

Shape deformation transfer methods operate directly on the surface either by using representations of 3D body shape that disentangle shape and pose~\cite{zhou2020unsupervised,cosmo2020limp} or by optimising mesh deformations~\cite{sumner2004deformation,baran2009semantic, zhou2010deformation,  boukhayma2017surface}. Some works~\cite{sumner2004deformation,baran2009semantic} consider motion retargeting as pose deformation transfer, encoding the pose of the source character and transferring the deformation of the associated mesh to the target. 
Other works~\cite{basset2021neural} consider motion retargeting as shape deformation transfer, encoding the shape identity of the target character and transferring it to the source character. A recent work adds short-term temporal context by considering $4$ consecutive frames~\cite{regateiro2022temporal}. These works showed impressive results by operating on meshes in correspondence. 
As providing spatio-temporal correspondences between source and target meshes is a difficult task, Wang~\etal~\cite{wang2020neural} propose a new style transfer approach that learns to transfer a character’s shape style onto another posed character while learning correspondences between point clouds with different structures. Chen~\etal~\cite{chen2021aniformer} extend this work to include short-term temporal information of $3$ frames. This improves results at the cost of requiring a target sequence instead of a single target frame as input during training.
Deformation-based approaches cannot be extended to learn long-term temporal context beyond a few consecutive frames due to computational complexity. We show experimentally that our method outperforms neural pose transfer (NPT)~\cite{wang2020neural}, the only existing correspondence-free method that can be trained without sequence-level supervision.

\subsection{Generic motion priors}

Generic motion priors representing dense human motion in a latent space that disentangles locomotion and dense body shape information have recently been applied to motion retargeting~\cite{jiang2022h4d,marsot22}. They allow to retarget the body motion to different body shape while considering long-term temporal context. Marsot~\etal~\cite{marsot22} use structured data and learn correlations between dense body shape and skeletal locomotion. Closer to our setting, H4D~\cite{jiang2022h4d} considers unstructured point clouds and retains dense surface details of the target shape. While motion priors account for long-term temporal context, they cannot easily be applied to sequences of arbitrary duration at inference as they operate on normalized temporal segments of motion. We demonstrate experimentally that our method outperforms H4D~\cite{jiang2022h4d}.

\subsection{Positioning}

Table~\ref{tab:rel} summarizes the positioning of our work~\wrt state-of-the-art. We classify approaches based on four criteria: using long-term temporal context ($0.5s$ or more) for training, allowing for online inference, modeling geometric detail, and operating on unstructured data for which no correspondences are known. By combining the advantages of skeleton-based retargeting and shape deformation models, we propose the first correspondence-free retargeting approach that learns long-term temporal context, allows for arbitrary duration at inference, and models geometric detail.

\begin{table}
\centering
{\small
\begin{tabular}{|p{2cm}|p{1cm}|p{1cm}|p{1cm}|p{1cm}|}
    \hline
    Method &  Tem\-poral context & Online inference & Geome\-tric detail  & Un\-struc\-tured \\
    \hline
    \hline
    \multicolumn{5}{|l|}{Skeleton-based} \\
    \hline
    \cite{aberman2020skeleton,villegas2018neural,lim2019pmnet,aberman2019learning} & \textcolor{Green}{$\checkmark$} & \textcolor{Green}{$\checkmark$} & \textcolor{Red}{\ding{55}} & \textcolor{Red}{\ding{55}} \\
    \cite{villegas21,zhang2023skinned} & \textcolor{Green}{$\checkmark$} & \textcolor{Green}{$\checkmark$} & \textcolor{Green}{$\checkmark$} & \textcolor{Red}{\ding{55}} \\
    \hline
    \hline
    \multicolumn{5}{|l|}{Shape deformation transfer} \\
    \hline
    \cite{chen2021intrinsic,basset2021neural,regateiro2022temporal} & \textcolor{Red}{\ding{55}} & \textcolor{Green}{$\checkmark$} & \textcolor{Green}{$\checkmark$} & \textcolor{Red}{\ding{55}}\\
    \cite{wang2020neural, chen2021aniformer}& \textcolor{Red}{\ding{55}} & \textcolor{Green}{$\checkmark$} & \textcolor{Green}{$\checkmark$} & \textcolor{Green}{$\checkmark$} \\
    \hline
    \hline
    \multicolumn{5}{|l|}{Motion priors} \\
    \hline
    \cite{marsot22} & \textcolor{Green}{$\checkmark$} & \textcolor{Red}{\ding{55}} & \textcolor{Green}{$\checkmark$} & \textcolor{Red}{\ding{55}} \\
    \cite{jiang2022h4d} & \textcolor{Green}{$\checkmark$} & \textcolor{Red}{\ding{55}} & \textcolor{Green}{$\checkmark$} & \textcolor{Green}{$\checkmark$} \\
    \hline
    \hline
    Ours& \textcolor{Green}{$\checkmark$} & \textcolor{Green}{$\checkmark$} & \textcolor{Green}{$\checkmark$} & \textcolor{Green}{$\checkmark$} \\
    \hline
\end{tabular}
}
\caption{Positioning~\wrt state-of-the-art retargeting approaches. We propose the first correspondence-free retargeting approach that learns long-term temporal context, allows for arbitrary duration at inference, all while modeling geometric detail at the surface level.}
\label{tab:rel}

\end{table}

\section{Motion retargeting method}

Our motion retargeting method takes as input a source motion of subject $A$ given as sequence of $n$ point clouds $\{\scan_i^A\}_{i=1}^{n}$ without temporal correspondences, and the target shape $B$ in T-pose given as point cloud, possibly with connectivity information, $\scan^B_{tpose}$. Our objective is to generate a sequence of retargeted scans $\{\scan_i^B\}_{i=1}^{n}$, where $\scan_i^B$ imitates the pose of $\scan_i^A$ while retaining the body shape of $\scan^B_{tpose}$, and is in correspondence with $\scan^B_{tpose}$. 

A common strategy for motion retargeting is to disentangle the high-level motion from the shape deformation caused by body shape, and to combine the source motion with the target body shape. This is computationally expensive when considering densely sampled input surfaces containing thousands of vertices per frame, especially if the surfaces are not in correspondence, which hinders existing approaches to learn long-term temporal context.

We overcome this challenge by making the hypothesis that source motion information is fully explained at a skeletal level, while geometric detail information is encoded in the target. This hypothesis leads to a retargeting objective consisting of two losses to be optimized during training. The first loss called source motion preservation loss $\motionloss$ encourages the retargeted motion to resemble $\{\scan_i^A\}_{i=1}^{n}$ at a skeletal level. Considering $\motionloss$ on a skeletal level has three major advantages. It provides correspondence information between input frames, drastically reduces the dimensionality of the data and fully encodes the shape induced variability in bone length. This allows to train using long-term context without running into complexity issues. The second loss called target geometry preservation loss $\geometryloss$ encourages each frame of the output sequence to have similar geometric details as $\scan^B_{tpose}$. Considering $\geometryloss$ independently per-frame allows processing high-dimensional point clouds that contain significant geometric detail. Our training procedure aims to optimize
\begin{equation}
	\mathcal{L} = \motionloss + \geometryloss.
\end{equation}

\begin{figure}[t]
    \centering
    \includegraphics[width=1\columnwidth]{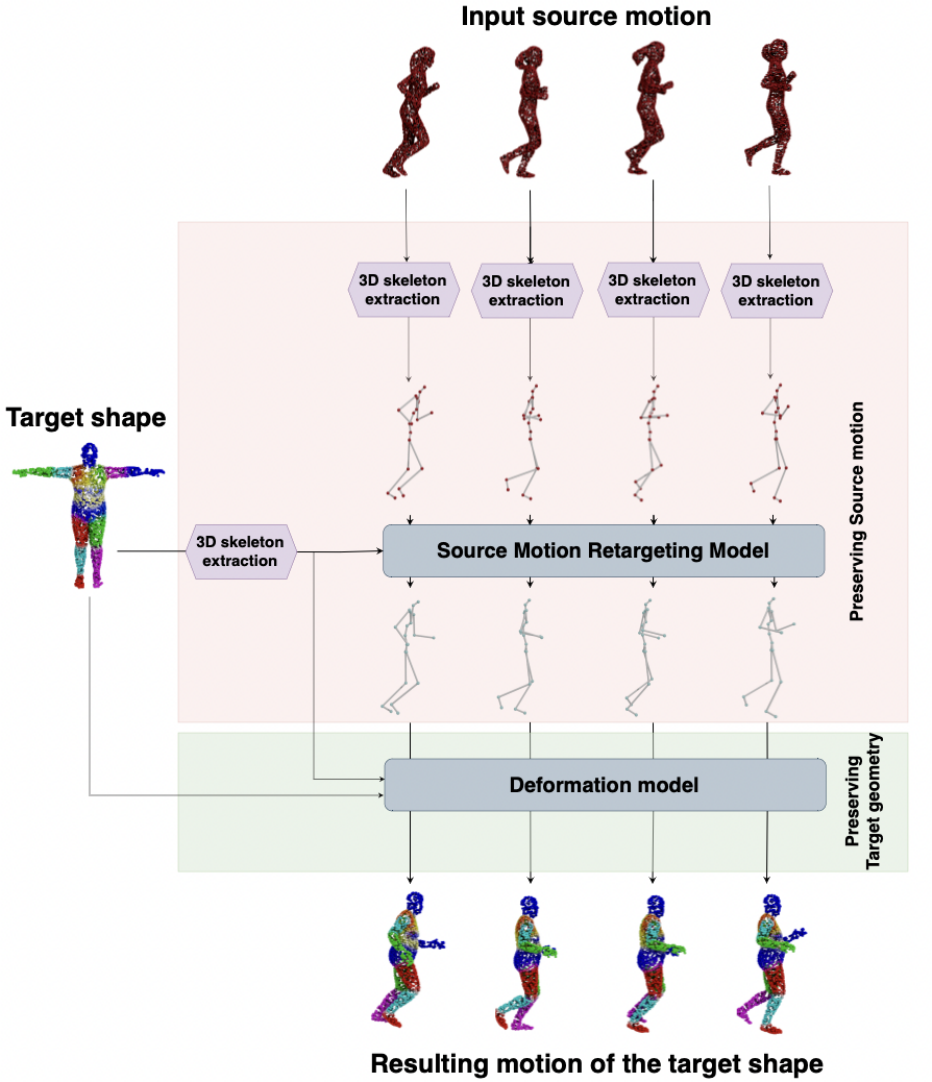}
    \caption{Our method takes a source sequence of unstructured point clouds and a target point cloud as input, and outputs the target character performing the input motion. The method first retargets the source motion to the target character at the skeletal level (red part), and then adds the target's surface details to the resulting motion using a deformation model (green part).}
    \label{fig:overview}
\end{figure}

During inference, our method then retargets the motion of $A$ to $B$ at the skeletal level and uses the resulting skeletal motion of $B$ to transfer geometric details from $\scan^B_{tpose}$ in a forward pass. This allows for efficient online retargeting. Fig.~\ref{fig:overview} gives a visual overview of our method. 

\subsection{Preserving source motion}

Our method aims to preserve the motion of the source character at a skeletal level. Working on the skeletal level is a key ingredient of our work, as this reduced representation allows to train using long-term temporal context without running into complexity issues. We first detail the architecture of our network and define the source motion preservation loss $\motionloss$, then outline how a skeletal representation is computed based on the inputs $\{\scan_i^A\}_{i=1}^{n}$ and $\scan^B_{tpose}$ of our method. In the following, $\{\joints^A_i\}_{i=1}^{n}$ and $\joints^B_{tpose}$ denote skeletal representations of $\{\scan_i^A\}_{i=1}^{n}$ and $\scan^B_{tpose}$.

\paragraph{Architecture} 

We consider the task of transferring the skeletal motion $\{\joints^A\}_{i=1}^{n}$ to $\joints^B_{tpose}$. Due to the lack of motion captures containing two different subjects performing the same motion, we train in an unsupervised way without using ground truth pairings between a motion and its retargeted version. To solve this problem, we were inspired by a skeletal motion retargeting model ($\retargetrnn$)~\cite{villegas2018neural} which given $\{\joints^A_i\}_{i=1}^{n}$ and $\joints^B_{tpose}$ outputs the relative joint rotations $\{\rotation^B_i\}_{i=1}^{n}$.

$\retargetrnn$ starts by extracting high level locomotion features from $\{\joints^A_{i}\}_{i=1}^{n}$ using a first recurrent network. These features are leveraged along with $\joints^B_{tpose}$ to generate joint rotations $\{\rotation^B_i\}_{i=1}^{n}$ using a second recurrent unit. The resulting rotations are applied to $\joints^B_{tpose}$ using a differentiable forward kinematics layer to generate the retargeted skeletal motion $\{\joints^B_i\}_{i=1}^{n}$.

\paragraph{Definition of $\motionloss$}

To train our network, we use cycle consistency losses, an adversarial loss, and a loss encouraging temporally smooth motions, all evaluated at the skeletal level. The cycle consistency losses transfer the motion from subject $A$ to $B$, and the resulting motion back to $A$, and compare the resulting motion of $A$ with its initial motion. Our loss is 
\begin{eqnarray}
    \motionloss &=& \mathcal{L}_{cycon} + \mathcal{L}_{adv} \nonumber \\
    &+& \lambda_{rot}\mathcal{L}_{rot}+ \lambda_{smooth}\mathcal{L}_{smooth}
\end{eqnarray}
with weights $\lambda_{rot}=0.01$ and $\lambda_{smooth}=0.1$. 

This loss is inspired by $\retargetrnn$, which uses cycle consistency loss $\mathcal{L}_{cycon} = \text{MSE}(\hat\joints^A_i,\joints^A_i)$, where MSE is the mean squared error, on joint positions and adversarial loss $\mathcal{L}_{adv}$. $\mathcal{L}_{adv}$ leverages a discriminator network which is trained in a min-max game with the retargeting network to differentiate between real and retargeted motions.
To improve performance, we add two additional loss functions to regularize the generated motions. Our first loss is a cycle consistency loss on the joint rotations, which allows to prevent unrealistic motions, and can be written as $\mathcal{L}_{rot} = \text{MSE}(\hat\rotation^A_i,\rotation^A_i)$. Rather than representing joint rotations using quaternions, we represent them in 6D as this representation was shown to be beneficial for deep learning applications~\cite{zhou2019continuity}. The second loss allows for temporal smoothing of the motions by aiming to reconstruct velocities, and can be written as $\mathcal{L}_{smooth} = \text{MSE}(\velocity\hat\joints^A_i,\velocity\joints^A_i)$, with $\velocity\joints_i =  \joints_{i+1} - \joints_i$. 

\paragraph{3D skeleton extraction}

It remains to outline how skeletons $\{\joints^A\}_{i=1}^{n}$ and $\joints^B_{tpose}$ are extracted from $\{\scan_i^A\}_{i=1}^{n}$ and $\scan^B_{tpose}$. To allow the processing of arbitrarily long input sequences, the skeleton extraction operates per-frame. To operate on unstructured point clouds, the skeleton extraction needs to be robust~\wrt the number and order of the observed points. To achieve this, we use the PointFormer~\cite{zhao2021point} architecture to extract order-invariant features from the point cloud, followed by a multi layer perceptron to regress the joint positions from these features. We choose to use the PointFormer architecture as it extracts local features of the point cloud allowing for precise joint predictions and good generalization to unseen poses. We denote this function, which recovers parameterized skeletons by $\joints = \skeletonextract(\scan)$. The skeleton extraction is trained in a supervised way while minimizing the MSE between the ground truth joints $\joints$ and the predicted ones as $\mathcal{L}_{\skeletonextract} = \text{MSE}(\skeletonextract(\scan),\joints)$. Our method is independent of the skeleton parameterization that is chosen. In our implementation, the skeleton is parameterized by 22 joint positions corresponding to all joints except the root joint of the SMPL body model~\cite{loper2015smpl}. 

\subsection{Preserving target geometry}

Our method aims to preserve the detailed geometry of the target character at a surface level. We achieve this by learning a spatially implicit geometry aware deformation model allowing to repose target character $\scan^B_{tpose}$ into an arbitrary pose $\rotation^B$ defined on skeleton $\joints^B_{tpose}$. Defining an implicit model offers the key advantage of allowing for arbitrarily sampled input during inference. We outline the deformation model then detail our network architecture and define the target geometry preservation loss $\geometryloss$.

\paragraph{Deformation model}

A common strategy to animate 3D meshes given their skeletal motion is to associate surface vertices to the skeleton joints by a set of skinning weights and to animate the shapes using skinning techniques. While recent works in this area show impressive results~\cite{ouyang2020autoskin,xu2020rignet,yang2021s3,chen2021snarf,mosella2022skinningnet}, they are typically applicable to registered data or supervised with skinning weights during training, which are not given in our case.

To allow for arbitrarily sampled input point clouds, we learn a geometry-aware deformation model that can be evaluated for any surface point of a human in T-pose. Inspired by works that build human shape spaces~\cite{allen2006learning,loper2015smpl}, we model deformations using two parts: a skinning weights predictor and a pose-corrective offset predictor. The pose corrective offset is added to vertices in T-pose before applying linear blend skinning to diminish artifacts. We choose this model for its simplicity and for its excellent performance when modeling human deformations. 

Given the target joints $\joints^B_{tpose}$, the set of rotations $\{\rotation^B_i\}_{i=1}^{n}$, the skinning weights $\weights^B_j$ and the pose corrective offsets $\{\blpose^B_{i,j}\}_{i=1}^{n}$ of point $p_j$ of $\scan^B_{tpose}$, we denote the function to generate the $j$-th vertex of frame $i$ by
\begin{equation}
\scan_{i,j}^B = \text{retarget}\left(\rotation^B_i,\joints_{tpose}^B, \weights^B_j,p_j+\blpose^B_{i,j}\right).
\end{equation}

\paragraph{Architecture} 

Our architecture consists of two networks. First, a skinning weights predictor network $\weightsnet$ that takes the target joints $\joints^B_{tpose}$ and a 3D point $p$ of $\scan^B_{tpose}$ as input and outputs one skinning weight per joint $\weights = \weightsnet(\joints^B_{tpose}, p)$. Inspired by~\cite{chen2021snarf}, we constrain the predicted skinning weights to sum to 1 using a softmax activation. Second, a pose-corrective offsets network $\blposenet$ takes relative joint rotations $\rotation^B$ and $p$ as input and outputs an offset vector in $\mathbb{R}^3$ as $\blpose^B = \blposenet(\rotation^B, p)$. The networks are modeled by two three layer MLPs. The first MLP models $\weightsnet$ and is followed by a softmax activation. The second MLP models $\blposenet$.

\paragraph{Definition of $\geometryloss$}

Our networks are trained on pairs of frames of the same identity with known poses $\scan_{tpose}$ and $\scan_{\rotation}$ with joint rotations $\rotation$. This allows for a point to point reconstruction loss. Consider point $p$ of $\scan_{tpose}$ and its corresponding point $p'$ of $\scan_{\rotation}$. Our loss can be written as
\begin{equation}
	\geometryloss =  \underset{ (p,p') }{  \sum} || p' - \text{retarget}(\rotation,\joints_{tpose},\weights, p+\blpose)||^2,
\end{equation}
where $\weights = \weightsnet (\joints_{tpose}, p)$ and $\blpose = \blposenet(\rotation, p)$. 

\subsection{Training}

Inspired by recent works~\cite{jiang2022h4d,jiang2020bcnet,mehta2020xnect,zhang2019predicting}, we choose a stage-wise training strategy to improve stability and reduce computational complexity. This is done by optimizing for $\motionloss$ and $\geometryloss$ in two independent stages using skeletal motions for $\motionloss$ and densely sampled static point clouds for $\geometryloss$.

For training, we leverage the AMASS dataset~\cite{mahmood2019amass}, which contains a collection of motion capture datasets that have been fitted by a parametric body model~\cite{loper2015smpl} to obtain dense per-frame representations. For each frame, aligned surface and skeleton information is given. All the motion sequences are temporally aligned at 30 frames per second (FPS). As training data, we consider a subset of 120 body shapes performing 2536 different motions for a total of 65000 seconds of motion. As validation data, we consider a subset of 3 body shapes performing 24 different motions for a total of 147 seconds of motion and as test set we consider 7 body shapes performing 44 motions for a total of 1715 second of motion. We call this test set \emph{AMASS test set}. 

All networks handling geometrically detailed data are trained by sampling one frame for each second of motion. To avoid learning the topology bias of a template mesh, we randomly uniformly sample $N$ points on the surface of each 3D mesh and add Gaussian noise to generate input scans.

To train the motion retargeting at the skeletal level, we randomly sample sequences of fixed duration from AMASS and consider their skeletal motion only. The 3D skeleton extractor is trained using static 3D point clouds sampled from AMASS with their corresponding skeletons. 

The geometric deformation model is trained using static AMASS data by considering pairs of frames $\{\scan_{tpose},\scan_{\rotation}\}$ of a same person in correspondence and with known rotations. To preserve correspondences while removing the bias due to SMPL topology, the models are uniformly resampled while preserving correspondence information.

\section{Experiments}

We now evaluate our main contributions: that learning from long-term temporal context improves the results, that our method outperforms state-of-the-art, and generalizes to previously unseen target shapes and source motions. 
We first show quantitatively that considering long-term temporal context improves the accuracy of motion retargeting. Second, we present quantitative comparisons using both standard error measures and evaluations via user studies to state-of-the-art results, considering both geometric detail preservation and skeletal-level motion retargeting. We consider challenging shape transfers with both naked and clothed target shapes where both shapes and motions are unseen during training. Finally, we show results that retarget the raw output of a 4D multi-view acquisition platform to new target characters. More results are in supplementary.

\paragraph{Data}
 
For a fair evaluation of the different methods considered in our comparison, we evaluate all methods on two test sets. First, a test set of representative naked human body shapes performing the same set of long-term motions. To build this dataset, we consider 4 body shapes created using the SMPL model~\cite{loper2015smpl}, as is commonly done when evaluating human deformation transfer methods~\eg~\cite{regateiro2022temporal}. We sample body shapes at $\pm 2$ standard deviations along the first 2 principal components to cover the main variabilities of human body shape, namely female, male, skinny and corpulent. Skeleton-based retargeting methods~\eg~\cite{villegas21} commonly evaluate on the Mixamo dataset~\footnote[3]{https://www.mixamo.com}. Inspired by this, we create and retarget a set of 4 motions to all body shapes using Mixamo to generate corresponding ground truth motions. We call this \emph{SMPL test set}. The second test set considers characters with clothing performing long-term motions. This test set allows to evaluate the generalization of different methods to geometric detail in the target shape. To generate this test set, we take 4 characters with tight clothes from CAPE~\cite{ma2020learning}, namely using the outfits called ``short long'', ``short short'', ``long short'' and ``long long'' in CAPE, and retarget 4 motions from the Mixamo dataset to each of these models. We call this \emph{CAPE test set}. Note that for both of SMPL and CAPE test sets, the 4 motions, namely ``Walking'', ``Jogging'', ``Throw'' and ``Hip Hop Dancing'', are varied in terms of global trajectory and local motions and none of the body shapes or motions were observed by any of the methods during training. Furthermore, by using Mixamo to retarget the same motion to different body shapes, ground truth retargeting results are available for quantitative testing. We also use a test set of untracked data acquired using a multi-view camera setup~\cite{marsot22}. For this dataset, no ground truth retargeting is available, and we provide qualitative results. We call this \emph{multi-view test set}.

\paragraph{Evaluation metrics}

The goal is to evaluate the retargeting results in terms of the overall preservation of the motion and the detail-preservation of the target geometry. 

To evaluate motion, we use two complementary metrics that operate exclusively on the skeletal level. First, we consider the mean-per-joint error (MPJPE) between the ground truth and the retargeting result, which evaluates the accuracy of the joint positions, and the Procrustes aligned MPJPE (PA-MPJPE), which eliminates the error in global displacement. Second, to evaluate motion smoothness, we consider the mean acceleration difference between ground truth predicted motions (Acc) and its Procrustes aligned (PA-Acc) version. 

To evaluate detail-preservation of the target geometry, we use two complementary metrics that operate on the surface. The first are mean-per-vertex distance (MPVD) and Procrustes aligned MPVD (PA-MPVD) between the ground truth and the retargeting result, which evaluate the global extrinsic accuracy of the predicted surface. Second, to evaluate the preservation of intrinsic geometry, we compute a mean difference in edge length (MDEL) between the ground truth and the retargeting result. As we operate on point clouds, we create edges by connecting the 6 closest neighbors of every point in the ground truth. 

\subsection{Learning with long-term temporal context}

Our first experiment demonstrates that considering temporal context beyond a few frames during training is beneficial to motion retargeting. We train our model with motion sequences containing different numbers of frames,~\ie for each model, all training sequences have a fixed number of frames, which ranges from $5$ frames (similar to shape deformation transfer methods~\cite{chen2021aniformer,regateiro2022temporal}) to $60$ frames (similar to skeleton based methods~\cite{villegas2018neural,lim2019pmnet,aberman2019learning}). Table~\ref{tab:quantitative-comp} shows the results. Including long-term context improves almost all metrics up to $30$ frames. In all following experiments, we use the model trained with sequences of $30$ frames.

\begin{table}[b]
    \centering
    \scalebox{0.85}{\begin{tabular}{|@{\thinspace}p{2.2cm}@{\thinspace}||@{\thinspace}p{0.95cm}@{\thinspace}|@{\thinspace}p{0.95cm}@{\thinspace}|@{\thinspace}p{0.95cm}@{\thinspace}|@{\thinspace}p{0.95cm}@{\thinspace}||@{\thinspace}p{0.95cm}@{\thinspace}|@{\thinspace}p{0.95cm}@{\thinspace}|@{\thinspace}p{0.95cm}@{\thinspace}|}
    \hline
     & \multicolumn{4}{@{\thinspace}c@{\thinspace}||@{\thinspace}}{Skeletal motion} &  \multicolumn{3}{c|}{Detail preserv.} \\
     \hline
	  \small{Context duration} &  \small{MPJPE (m)$\downarrow$}  & \small{PA-MPJPE (m)$\downarrow$} &  \small{Acc$\downarrow$} &  \small{PA-Acc$\downarrow$} &  \small{MPVD (m)$\downarrow$}  &  \small{PA-MPVD (m)$\downarrow$} &  \small{MEDL (mm)$\downarrow$}\\
	 \hline
	 \hline
	  \small{0.16s (5 frames)} &0.203 & 0.073 & 0.021 & 0.009 & 0.158 & 0.061 & \textbf{0.505}  \\
	 \hline
	  \small{0.33s (10 frames)} & 0.167 & 0.050 & 0.021 & 0.008 & 0.137 & 0.045 & 0.507     \\
	 \hline
	  \small{0.5s(15 frames)} & 0.161 & 0.046 & \textbf{0.020} & 0.008 & 0.136 & 0.044 & 0.519  \\
	 \hline
	  \small{1s (30 frames)} & \textbf{0.158} &  \textbf{0.043}  & 0.021 & \textbf{0.008} & 0.134 & \textbf{0.040} & 0.524 \\
	 \hline                   
	  \small{2s (60 frames)} & 0.160 & 0.044 & 0.023 & 0.008 & \textbf{0.131 }& 0.041 & 0.509 \\  
	 \hline
	 \end{tabular}}
    \caption{Learning with different temporal contexts on SMPL test set. Training with long-term context of $1s$ improves the results.}

   \label{tab:quantitative-comp}
\end{table}

\subsection{Quantitative comparison to state-of-the-art} 

\begin{table}
    \centering
    {\small
    \scalebox{0.95}{\begin{tabular}{|@{\thinspace}p{1.3cm}@{\thinspace}||@{\thinspace}p{0.95cm}@{\thinspace}|@{\thinspace}p{0.95cm}@{\thinspace}|@{\thinspace}p{0.80cm}@{\thinspace}|@{\thinspace}p{0.80cm}@{\thinspace}||@{\thinspace}p{0.95cm}@{\thinspace}|@{\thinspace}p{0.95cm}@{\thinspace}|@{\thinspace}p{1cm}@{\thinspace}|}
    \hline
     & \multicolumn{4}{@{\thinspace}c@{\thinspace}||@{\thinspace}}{Skeletal motion} &  \multicolumn{3}{c|}{Detail preserv.} \\
     \hline
	 & MPJPE (m) $\downarrow$ & PA-MPJPE (m) $\downarrow$ & Acc $\downarrow$ & PA-Acc $\downarrow$ & MPVD (m) $\downarrow$ & PA-MPVD (m) $\downarrow$ & MDEL (mm) $\downarrow$ \\
	 \hline
	\multicolumn{8}{l}{\textbf{Naked target shapes from SMPL test set}} \\
	\hline
	 H4D~\cite{jiang2022h4d} & 0.238 & 0.096 & \textbf{0.019} & 0.014 & 0.152 & 0.078 & 402.238\\
	 \hline
	 NPT~\cite{wang2020neural} &0.388 & 0.165 & 0.024 & 0.014 & 0.227 & 0.132 & 2.739   \\  
	 \hline
	 Ours &\textbf{0.158} & \textbf{0.043} & 0.021 & \textbf{0.008} & \textbf{0.134} & \textbf{0.040} & \textbf{0.524}\\
	 \hline
	 \multicolumn{8}{l}{\textbf{Clothed target shapes from CAPE test set}} \\
	 \hline
	 H4D~\cite{jiang2022h4d} & 0.161 & 0.091 & \textbf{0.019} & 0.014 & 0.096 & 0.074 & 395.683\\
	 \hline
	 NPT~\cite{wang2020neural} & 0.373 & 0.168 & 0.022 & 0.013 & 0.173 & 0.135 & 2.845 \\  
	 \hline
	  Ours &\textbf{0.105} & \textbf{0.040} & 0.023 & \textbf{0.008} & \textbf{0.075} & \textbf{0.038} & \textbf{0.539} \\
	 \hline
	 \end{tabular}}
	 }
    \caption{Comparison to state-of-the-art on naked (top) and clothed (bottom) target shapes. Best performing scores shown in bold.}

   \label{tab:quantitative-comp1}
\end{table}

We now present a comparative analysis to state-of-the-art motion retargeting methods. 

\textbf{Competing methods} As summarized in Table~\ref{tab:rel}, there are three lines of existing methods. Skeleton-based methods are not comparable to our approach as they require hand crafted skinning weights as input, which are not available for our test sets. 
We therefore compare our method to correspondence-free deformation transfer methods and motion priors. For deformation transfer methods, we compare to NPT~\cite{wang2020neural}. Aniformers~\cite{chen2021aniformer} requires ground truth pairings of different individuals performing the same motion during training, which are not available in our case, making comparison impossible. For motion priors, we compare to H4D~\cite{jiang2022h4d}. We provide a full identity sequence to this method to extract body shape parameters.

\textbf{Quantitative results}
Table~\ref{tab:quantitative-comp1} provides quantitative results when considering shapes of the SMPL and CAPE test sets.
Our method significantly outperforms NPT on almost all metrics on both datasets. The reason is that NPT operates per frame without any temporal context and sometimes leads to results that are temporally inconsistent.
Our method also outperforms H4D on almost all evaluation metrics on both datasets. In particular, the skeletal joint positions after Procrustes alignment are significantly more accurate for both test sets, and without Procrustes alignment, the mean is $4.9cm$ more accurate for naked target shapes, while being within $2mm$ for clothed ones. Joint accelerations are more accurate when using our model. Geometric detail is significantly better preserved using our model when considering Procrustes alignment for both datasets, and without Procrustes alignment, the errors of both models are similar. This implies that our model retains geometric detail better. 

\textbf{Comparison to retargeting method using correspondences} As we outperform correspondence free retargeting methods by a large margin, we also compare our method to TST~\cite{regateiro2022temporal}, a state-of-the-art deformation transfer method considering short-term temporal dynamics that requires point-to-point correspondences at training and inference, and provide this method with correspondences. Our method only performs slightly worse than TST on the evaluation metrics, with all errors being within $2cm$ of TST. This performance, close to a method leveraging correspondences, highlights the potential of our correspondence-free method. Detailed numerical results of evaluation metrics are provided in supplementary material.

\subsection{User Study}
We further evaluate how our retargeting results are perceived by humans \wrt both preservation of motion and geometric detail by conducting a user study. 

\textbf{Design} We designed a within-subject perceptual experiment with retargeting methods as a main factor. Since NPT performs significantly worse than competing methods quantitatively, we did not include it in the perceptual evaluation in order to minimize the duration of the experiment for each user. We compared the results of our method to H4D and TST, which uses correspondences. 30 unpaid volunteers participated in this study. 
For each trial, they were presented stimuli showing a source motion, a target shape, and a retargeting result and were asked to rate motion and shape similarity respectively with source motion and target shape on a 7-point Likert scale. 15 repetitions were performed for each method using 3 types of motion (walking, hip-hop, throwing) and 5 different shape transfers. Participants performed 45 trials, presented in a randomized order. Statistical analysis was done using ART-ANOVAs and post-hoc Tukey tests. 

\textbf{Results}
Results highlight a main effect of the method both on motion and shape components (p$<$0.0001). Post-hoc analysis shows that users rated our method to preserve motion and body shape significantly better than H4D (p$<$0.0001). TST outperforms our method on motion preservation (p$<$0.001), while our method better preserves the body shape of the target (p$<$0.0001). More details about the design and results are provided in supplementary.

\begin{figure*}
    \centering
    \includegraphics[width=\textwidth,trim={0 47pt 0 0},clip]{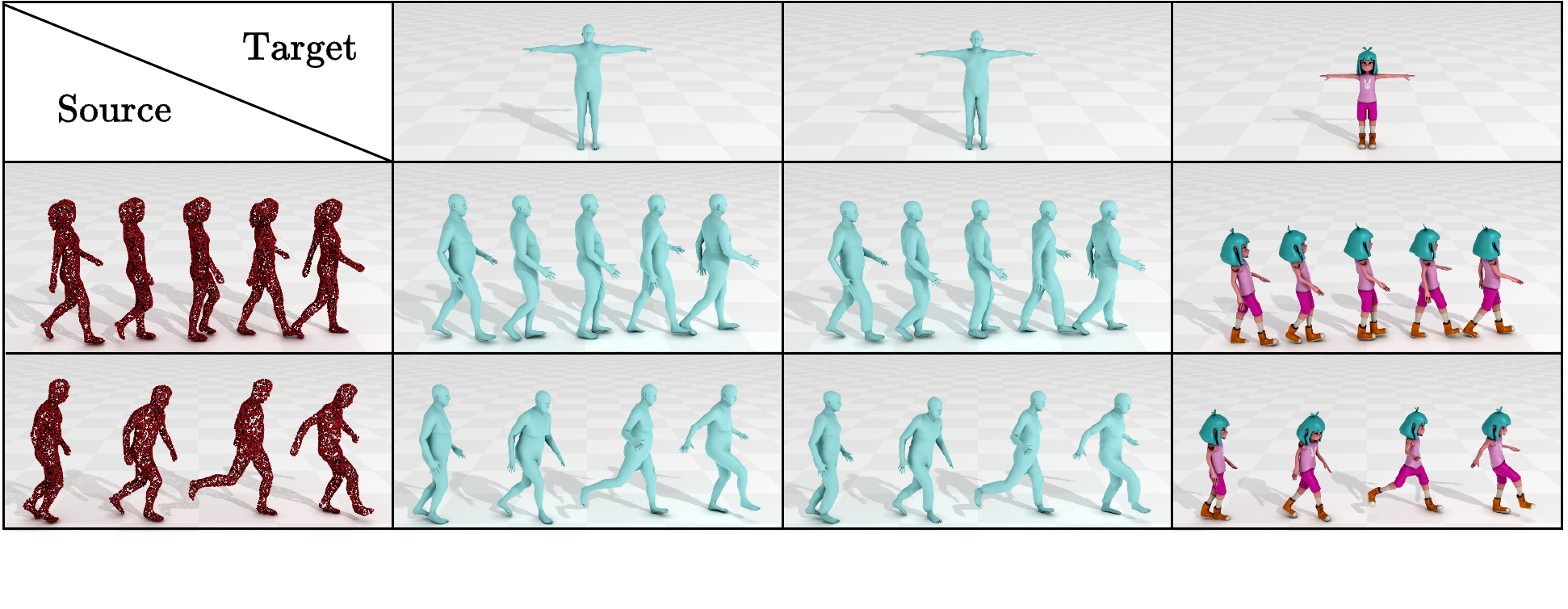}
    \caption{Animating target shapes with untracked captured 4D data directly. We consider a walking motion (top) and a kicking motion (bottom), which are retargeted to a naked (left),clothed (middle) and a CAD-generated (right) target shape.}
    \label{fig:fig-raw-retargeting}
\end{figure*}

\subsection{Ablations}

We provide ablations for the architectures optimizing $\motionloss$ and $\geometryloss$. 

\textbf{Source motion preservation} We first evaluate motion transfer on the skeletal level on the densely sampled geometry of the SMPL test set. Table~\ref{tab:abl_rot} shows that using the 6D rotation representation and the rotation supervision using $\mathcal{L}_{rot}$ improve the retargeting. The transition from quaternion to 6D rotation leads to a significant improvement on almost all metrics. Supervising with $\mathcal{L}_{rot}$ also leads to an improvement in the geometric detail preservation around the feet, wrists and head joints. This is because these joints are leaves of the hierarchical skeleton, so they have no influence on $\mathcal{L}_{cycon}$ but they do influence $\mathcal{L}_{rot}$. 

\begin{table}[h]
  \centering
   {\small\scalebox{0.95}{\begin{tabular}{|@{\thinspace}p{1.9cm}@{\thinspace}||@{\thinspace}p{0.91cm}@{\thinspace}|@{\thinspace}p{0.91cm}@{\thinspace}|@{\thinspace}p{0.72cm}@{\thinspace}|@{\thinspace}p{0.72cm}@{\thinspace}||@{\thinspace}p{0.91cm}@{\thinspace}|@{\thinspace}p{0.91cm}@{\thinspace}|@{\thinspace}p{0.90cm}@{\thinspace}|}
   \hline
     & \multicolumn{4}{@{\thinspace}c@{\thinspace}||@{\thinspace}}{Skeletal motion} &  \multicolumn{3}{c|}{Detail preserv.} \\
   \hline
	  & MPJPE (m) $\downarrow$ & PA-MPJPE (m) $\downarrow$ & Acc $\downarrow$ & PA-Acc $\downarrow$ & MPVD (m) $\downarrow$ & PA-MPVD (m) $\downarrow$ & MEDL (m) $\downarrow$\\
	  \hline
	 \cite{villegas2018neural} & 0.163 & 0.056 & 0.023 & 0.011 & 0.165 & 0.089 & 1.597\\
 \hline
	 \cite{villegas2018neural}+6D &0.159 & 0.043 & 0.024 & 0.008 & 0.137 & 0.048 & 0.684\\
	 \hline
	 \cite{villegas2018neural}+$\mathcal{L}_{rot}$ &0.161 & 0.044 & 0.021 & 0.008 & 0.156 & 0.076 & 1.445\\
	 \hline
	 \cite{villegas2018neural}+$\mathcal{L}_{rot}$+6D  &\textbf{0.158} &\textbf{0.043} &\textbf{0.021} &\textbf{0.008} & \textbf{0.134 }& \textbf{0.040} & \textbf{0.524} \\
	 \hline
	 \end{tabular}}}
   \caption{Ablation of motion retargeting. Quantitative evaluation on SMPL shapes with different rotation representation for $\rotation$ and ablation of the rotation cycle consistency loss $\mathcal{L}_{rot}$.}
  \label{tab:abl_rot}
\end{table}

To evaluate the 3D skeleton extraction, we use the AMASS test set. We compare between PointNet~\cite{qi2017pointnet}, which considers global features and PointFormer~\cite{zhao2021point}, which introduces local features. Using PointFormer improves the MPJPE to \textbf{21mm} over \textbf{46mm} for PointNet.

\textbf{Target geometry preservation}
To ablate the deformation model used to optimize $\geometryloss$, we use the AMASS test set. Using pose-corrective offsets improves precision by reducing the average reconstruction error from \textbf{8.4mm} to \textbf{5mm}. 
Further ablations are in supplementary.

\subsection{Animating target shape with captured 4D data}

We demonstrate our method's performance when animating a target shape with the raw 4D output of a multi-view acquisition platform. We retarget sequences of the multi-view test set directly to characters generated using SMPL, CAPE, and a Mixamo character designed using computer aided design (CAD) tools. Note that the input sequence suffers from acquisition noise and that no correspondence information is available,~\ie we input the untracked 4D sequence.

Fig.~\ref{fig:fig-raw-retargeting} shows the results obtained using our method. Note how the motion of the source sequence as well as the geometric detail of the different target shapes are preserved by our method. To the best of our knowledge, our method is the first that can retarget untracked 4D acquisition data online.

These examples show the robustness of our method to unseen shapes. The first source motion exhibits a body shape with hair, not seen during training, demonstrating the robustness of our method to unseen source shapes. The preservation of the geometry for CAPE and CAD generated target shapes also demonstrates that our model generalizes well when considering unseen target shapes.   

Quantitative and qualitative results show our method to generalise well on unseen motions \eg from Mixamo and unseen shapes \eg clothed shapes from CAPE and untracked 4D output of multi-view acquisition platforms. 

\subsection{Limitations} 
While our method gives state-of-the-art results for the correspondence-free motion retargeting problem, limitations remain. It cannot generalize on clothed shapes with wide garments due to limitations resulting from our deformation model. The method is also restricted to shapes that can be parameterized by a skeleton with fixed topology and is not able to capture detailed hand or facial motion. Potential negative societal impact is discussed in supplementary.

\section{Conclusion}

We proposed the first online retargeting method that allows to animate a target shape with a correspondence-free source motion. We demonstrated that including long term temporal context of $1s$ is beneficial when retargeting dense motion. Our low dimensional intermediate skeletal representation combined with the skinning prior generalizes well to unseeen shapes and motions. In particular, we demonstrate that our model, learned exclusively on naked body shapes, generalizes to inputs with hair and clothing.

Future works include allowing for extensions to complex garments such as wide or layered clothing. One option is to explicitly include clothing in the model. Extending the solution to hands and expressions is also interesting.

\section{Acknowledgement}

We thank Jo\~{a}o Regateiro and Edmond Boyer for helpful discussions. This work was funded by the French National Research Agency (ANR) 3DMOVE - 19-CE23-0013.

{
   \small
   \bibliographystyle{ieeenat_fullname}
  \bibliography{main}
}

\clearpage
\maketitlesupplementary

\appendix

This supplementary material contains the network architecture and implementation details of our proposed framework and more details about the ablations. We further show comparisons of our method to a state-of-art retargeting method that uses correspondences, and provide additional explanations of the user study we conducted.

\section{Network architecture and implementation details}

Our method aims to preserve the motion of the source character at the skeletal level via a source motion retargeting model, and the detailed geometry of the target shape using a deformation model. The following provides details for each of the two parts.

In the following, $N$ is the batch size, $V$ is the number of vertices of scan  $\scan$, and $J$ is the number of joints of skeleton $\joints$.

All parts are implemented in PyTorch and optimized using Adam~\cite{kingma2014adam}. The training of the whole framework takes approximately one day on a GeForce RTX 2080 TI. The whole model has 14.6 million trainable parameters. The source motion retargeting model has 5 793 927 parameters, the 3D skeleton extraction model has 9 585 730 parameters and the deformation model has 257 564 parameters. 

\subsection{Source motion preservation}
The source motion retargeting model is a reimplementation of ~\cite{villegas2018neural}. We summarize the architecture in Table~\ref{retarg}. FK denotes a forward kinematic layer.
\begin{table}[h]
    \centering
    \scalebox{0.8}{
    \begin{tabular}{|c|c|c|c|c|}
    
     \hline
	   Index & Inputs & Operation & Output shape\\
	   \hline
	   (1) & Input & Joint positions {$\joints$} & $N \times 3 \times J$ \\
	   \hline
	   (1') & Input & Target skeleton $\joints^B_{tpose}$ & $N \times 3 \times J$ \\
	   \hline
	   (2) & Input & Translations & $N \times 3$ \\
	   
	 \hline
	  (3) & (1) + (2) & GRU ($3\times (J+1) \rightarrow 512, 2$) &  $N \times 512 \times J$ \\
	  \  & \ & Dropout ($p=0.2$)  & \\
	 \hline
	  (4) & (3) + (1') & GRU ($3 \times J+512\rightarrow 512, 2$) & $N \times 512 $  \\
	  \  & \ & Dropout ($p=0.2$) &  \\
	  \hline
	   (5) &  (4) & Linear $(512 \rightarrow 6 \times J)$ &   $N \times 6 \times J$ \\
	   \hline
	   (6) &  (4) & Linear $(512 \rightarrow 3)$ & $N \times 3 $   \\
	   \hline
	   (7) &  (5) + (1') & FK & $N \times 3 \times J$ \\
% 	   \hline
% 	  (1')  & (1) & Feature-Transformation  & $N \times 3 \times V$ \\
	 \hline
	 \end{tabular}}
    \caption{The network architecture for source motion retargeting.}
   \label{retarg}
\end{table}

The network is trained for 450 epochs where each epoch sees 2 316 examples. The learning rate start at 1e-4 and is divided by 10 every 150 epochs. 

\paragraph{3D skeleton extraction}
Table~\ref{skreg} provides details on the architecture of the 3D skeleton extraction module.
\begin{table}[h]
   \centering
    \scalebox{0.8}{
    \begin{tabular}{|c|c|c|c|c|}
     \hline
	   Index & Inputs & Operation & Output shape & Activation\\
	 \hline
	  (1)  & Input & Source scan $\scan^A$ & $N \times 3 \times V$ & -\\
	  \hline
	  (1')  & (1) & backbone layer \cite{zhao2021point} & $N \times 4 \times512$ & -\\
	  \hline
	  (2)  & (1') & Mean & $N \times512$ & -\\
	  \hline
	  (3)  & (2) & Linear $(512 \rightarrow 256)$  & $N \times 256$  & Relu\\
	  \hline
	  (4)  & (3) & Linear $(256 \rightarrow 64)$ & $N \times 64$  & Relu\\
	  \hline
	  (5)  & (4) & Linear ($64 \rightarrow J\times3$)  & $N \times 3 \times J$ & -\\
	  
	  \hline
	 \end{tabular}}
    \caption{The network architecture for 3D skeleton extraction.
    }
   \label{skreg}
\end{table}

The network is trained for 34 epochs where each epoch sees 65 000 examples. The learning rate is 1e-4 for the first 30 epochs then 1e-5 for 2 epochs and 1e-6 for 2  epochs.

\subsection{Target geometry preservation}
The architecture of the deformation module is composed of the skinning weights predictor network detailed in Table~\ref{skinning2} and the pose corrective offset predictor network detailed in Table~\ref{skinning1}.

\begin{table}[h]
    \centering
    \scalebox{0.65}{
    \begin{tabular}{|c|c|c|c|c|c|}
    
     \hline
	   Index & Inputs & Operation & Output shape & Activation\\
	 \hline
	  (1) & \small{$N \times 3 \times (J +1) $} & Linear \small{$(3 \times (J +1)) \rightarrow 256)$} & $N \times 256$ & Relu  \\
	  \  & \ & Dropout \small{($p=0.2$)}& \ & \   \\
	 \hline
	 (2) & (1) & Linear \small{(256→256)} & $N \times 256$ & Relu \\
	  \  & \ & Dropout {($p=0.2$)}& \ & \   \\
	 \hline
	 (3) & (2) & Linear \small{$(256 \rightarrow J)$} & $N \times J$ & Softmax \\
	 \hline
	 \end{tabular}}
    \caption{The network architecture for skinning weights predictor.}
   \label{skinning2}
\end{table}

\begin{table}[h]
    \centering
    \scalebox{0.65}{
    \begin{tabular}{|c|c|c|c|c|c|}
    
     \hline
	   Index & Inputs & Operation & Output shape & Activation\\
	 \hline
	  (1) & \small{$N \times 3 \times (J +1) $} & Linear \small{$(3 \times (J+1) \rightarrow 256)$} & $N \times 256$ & Relu  \\
	  \  & \ & Dropout \small{($p=0.2$)}& \ & \   \\
	 \hline
	 (2) & (1) & Linear \small{(256→256)} & $N \times 256$ & Relu \\
	  \  & \ & Dropout {($p=0.2$)}& \ & \   \\
	 \hline
	 (3) & (2) & Linear \small{$(256 \rightarrow 3)$} & $N \times 3$ & - \\
	 \hline
	 \end{tabular}}
    \caption{The network architecture for the pose corrective offset predictor.}
   \label{skinning1}
\end{table}

The skinning weights predictor and pose corrective offset predictor networks are trained jointly for 28 epochs where each epoch sees 65 000 examples. For the first epoch, only the skinning weights predictor network is optimized. The learning rate is 1e-3.

\section{Comparison to retargeting method using correspondences, TST}
Table \ref{tab:corres} quantitatively compares our method to TST~\cite{regateiro2022temporal}, a state-of-the-art deformation transfer method that requires point-to-point correspondence. Note that while TST performs better than our method, the improvement is minor. This is a strong result showing that our correspondence-free method only gets slightly outperformed by a state-of-the-art method that relies on additional correspondence information.
\begin{table}[h]
    \centering
    {\small
    \scalebox{0.95}{\begin{tabular}{|@{\thinspace}p{1.3cm}@{\thinspace}||@{\thinspace}p{0.95cm}@{\thinspace}|@{\thinspace}p{0.95cm}@{\thinspace}|@{\thinspace}p{0.80cm}@{\thinspace}|@{\thinspace}p{0.80cm}@{\thinspace}||@{\thinspace}p{0.95cm}@{\thinspace}|@{\thinspace}p{0.95cm}@{\thinspace}|@{\thinspace}p{1cm}@{\thinspace}|}
    \hline
     & \multicolumn{4}{c||}{Skeletal motion} &  \multicolumn{3}{c|}{Detail preserv.} \\
     \hline
	 & MPJPE (m) $\downarrow$ & PA-MPJPE (m) $\downarrow$ & Acc $\downarrow$ & PA-Acc $\downarrow$ & MPVD (m) $\downarrow$ & PA-MPVD (m) $\downarrow$ & MDEL (mm) $\downarrow$ \\
	 \hline
	\multicolumn{8}{l}{\textbf{Naked target shapes from SMPL test set}} \\
	\hline

	 TST~\cite{regateiro2022temporal} & \textbf{0.152} & \textbf{0.028} & \textbf{0.005} & \textbf{0.004} & \textbf{0.130} & \textbf{0.028} & 0.866 \\

	 \hline
	 Ours &0.158 & 0.043 & 0.021 & 0.008 & 0.134 & 0.040 & \textbf{0.524}\\
	 \hline
	 \multicolumn{8}{l}{\textbf{Clothed target shapes from CAPE test set}} \\
	 
	 \hline

	 TST~\cite{regateiro2022temporal} & \textbf{0.096} & \textbf{0.027} & \textbf{0.005} & \textbf{0.004} & \textbf{0.064} & \textbf{0.028} & 0.953\\
	 \hline
	  Ours &0.105 & 0.040 & 0.023 & 0.008 & 0.075 & 0.038 & \textbf{0.539} \\
	 \hline
	 \end{tabular}}
	 }
    \caption{Comparison of our correspondence-free method to a state-of-the-art method which needs correspondences. The comparison is done on naked (top) and clothed (bottom) target shapes. }
   \label{tab:corres}
\end{table}

\section{Further ablation studies for source motion preservation }

We now present further ablation studies for the source motion retargeting model.

\begin{figure*}[h]
    \centering
    \includegraphics[width=0.8\textwidth]{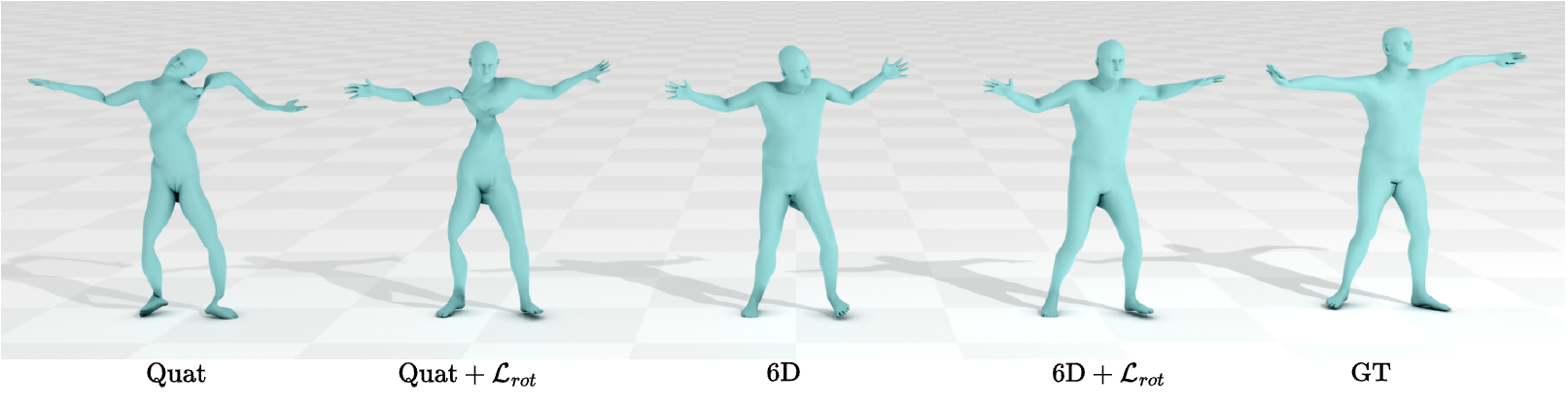}
    \caption{Ablation of {\retargetrnn}. Retargeting result on a challenging HipHop motion from a female to a male body shape. Quaternions are prone to twist, introducing $\mathcal{L}_{rot}$ improves the head and feet retargeting.}
    \label{fig:retargeting}
\end{figure*}

First, Table 4 in the main paper shows that using $\mathcal{L}_{rot}$ and the 6D representation lead to significant improvements on all quantitative metrics. Figure~\ref{fig:retargeting} further shows an example retargeting with different rotation representations and with and without the cycle consistency loss. Note that our changes significantly improve the retargeting result.

Second, we show quantitatively in Table~\ref{tab:abl_rot1} that adding $\mathcal{L}_{smooth}$ improves the source motion retargeting. We evaluate the skelatal motion since the $\mathcal{L}_{smooth}$ operates on skeletons.

\begin{table}[h]
    \centering
    {\small\scalebox{0.98}{\begin{tabular}{|@{\thinspace}p{1.9cm}@{\thinspace}||@{\thinspace}p{0.94cm}@{\thinspace}|@{\thinspace}p{0.94cm}@{\thinspace}|@{\thinspace}p{0.74cm}@{\thinspace}|@{\thinspace}p{0.74cm}@{\thinspace}|}
    \hline
     & \multicolumn{4}{@{\thinspace}c@{\thinspace}|}{Skeletal motion}\\
    \hline
	  & MPJPE (m) $\downarrow$ & PA-MPJPE (m) $\downarrow$ & Acc $\downarrow$ & PA-Acc $\downarrow$ \\
	  \hline

	 $\motionloss$  & 0.160 & 0.044 & 0.023 & 0.008 \\
	 without $\mathcal{L}_{smooth}$ & \ &  \ &  \ &  \\
	 \hline
	 $\motionloss$ &\textbf{0.158} &\textbf{0.043} &\textbf{0.021} &\textbf{0.008} \\
	  with $\mathcal{L}_{smooth}$ & \ &  \ &  \ &  \\
	 \hline
	 \end{tabular}}}
    \caption{Quantitative ablation of $\mathcal{L}_{smooth}$ on SMPL shapes.}
  \label{tab:abl_rot1}
\end{table}

\paragraph{3D skeleton extraction}

\begin{figure}[h]
    \centering
    \includegraphics[width=\columnwidth,trim={12cm 15cm 12cm 15cm},clip]{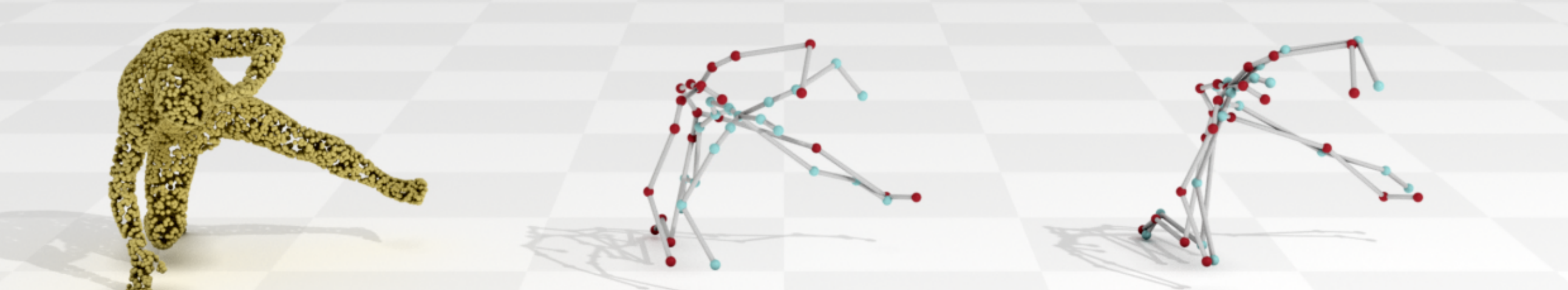}
    \caption{Ablation of 3D Skeleton extraction: with PointNet (middle) and PointFormer (right) on a challenging pose from the AMASS test set (right). Ground truth shown in red, regressed results in blue. }
    \label{fig:skr}
\end{figure}

In this paragraph, we study the effect of different point cloud encodings on the 3D skeleton extraction model. The main paper presents quantitative results in favour of using PointFormer over PointNet for this purpose. 
Figure \ref{fig:skr} shows a comparison on a challenging example. When using PointNet, the torso prediction is far from the ground truth while using PointFormer, the only noticeable errors are on the right foot and wrist joints.

% \begin{itemize}
   
% %\item tableau quantitative Pas besoin de tableau, numbers already in paper

% \end{itemize}
\section{User study}

\begin{figure*}[h]
    \centering
    \includegraphics[width=0.8\textwidth]{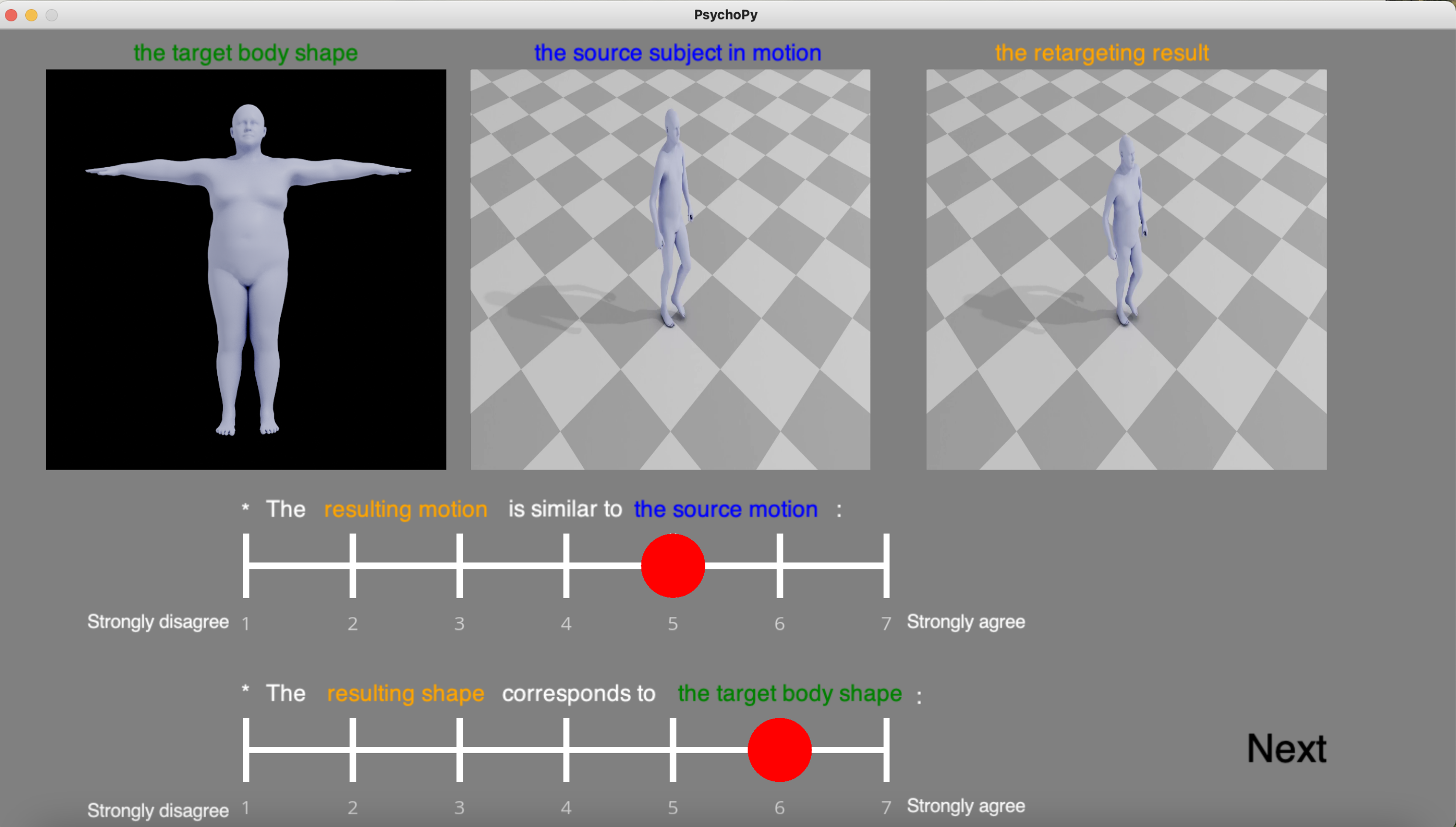}
    \caption{Screenshot of the stimuli and questions presented to participants during the user study.}
    \label{screenshot}
\end{figure*}

In order to further assess our method, we designed a user study to evaluate the perceived resulting motion with respect to the source motion, and the perceived resulting shape with respect to the target shape.

\subsection{Participants}
30 unpaid participants volunteered for the experiment (14 females, 16 males; age: average=32$\pm$11, range 21-64). They had different expertise levels in animation (median = 2, interquartile range (IQR) = 3, range 1-7 using a Likert-scale from 1 (novice) to 7 (expert)) and human motion (median = 3, IQR = 4, range 1-7 using a Likert-scale from 1 (novice) to 7 (expert)). They all had normal or corrected to-normal vision, and gave written and informed consent. The study conformed to the declaration of Helsinki, and was approved by our institutional ethics committee. 
%Participants were first asked to read and fill up a consent form. 

\subsection{Design}

For each trial, participants were seated in front of a 24" computer screen and were presented stimuli showing a source motion, a target shape, and a retargeting result, and asked to rate motion and shape similarity, respectively, with source motion and target shape on a 7-point Likert scale, as illustrated on Figure~\ref{screenshot}. 

\begin{figure*}
    \centering
    \includegraphics[width=0.6\textwidth]{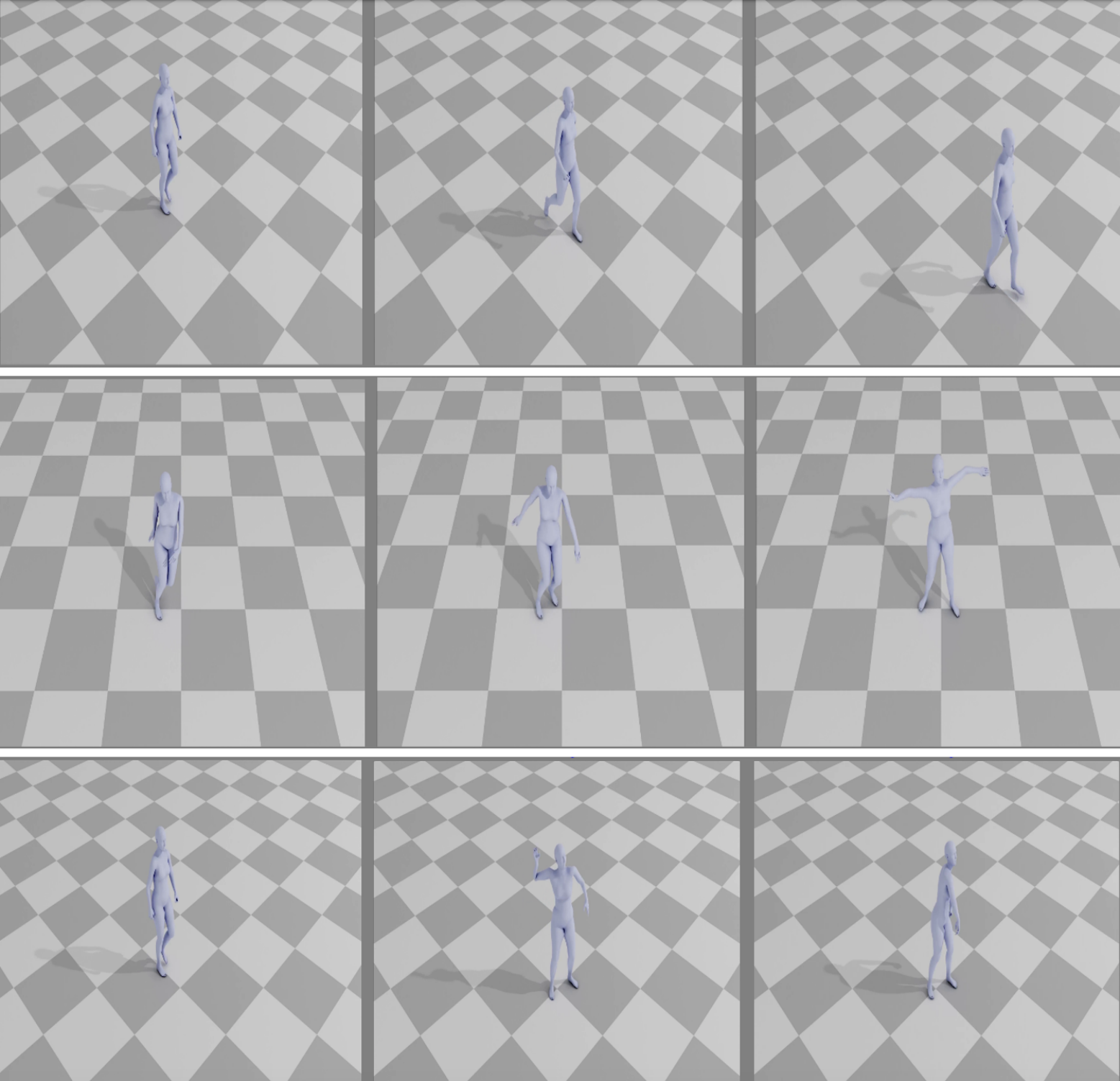}
    \caption{Illustration of the 3 motions selected for the user study. Top: Walking motion, middle: Hip-Hop motion and bottom: Throwing motion. Those examples were used as source motion for the female shape.} %were generated  using our method for transfer, when retargeting from female to male. }
    \label{motions}
\end{figure*}

We built the experiment using PsychoPy software. We used a within-subject design with retargeting methods as a main factor. We considered 3 retargeting methods: ours and H4D (without correspondences), and TST (with correspondences). 15 repetitions were performed for each method using different motions and shape transfers. We selected 3 types of motions, shown in Figure~\ref{motions}:  
\begin{enumerate}
    \item walking, which involves a global displacement with cyclic body-motions,
    \item throwing, which is a more local and discrete upper-body motion, and
    \item hip-hop, which is a complex motion.
\end{enumerate}

\begin{figure}[h]
    \centering
    \includegraphics[width=\columnwidth]{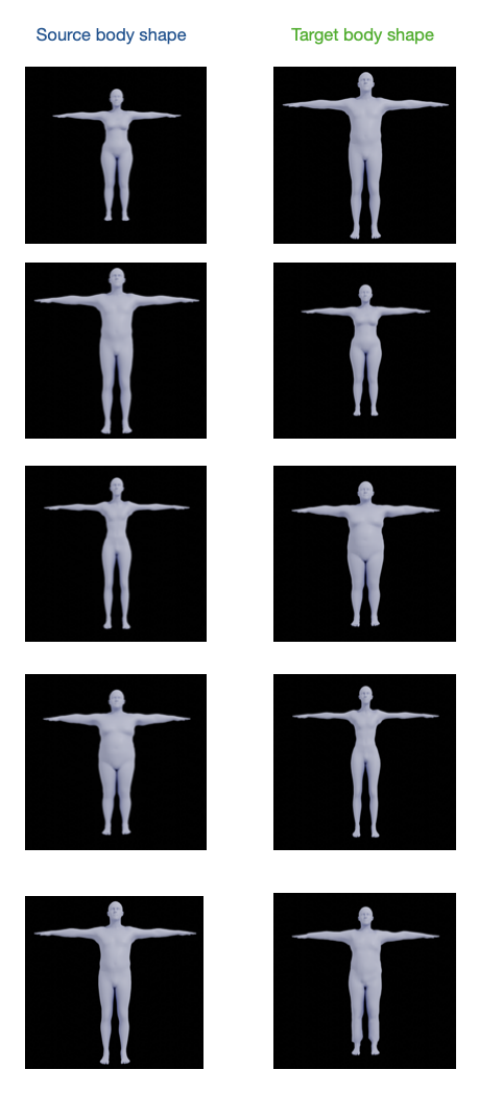}
    \caption{Illustration of the 5 shape transfers selected for the user study. Source shape is on the left, target shape is on the right.}
    \label{shapes}
\end{figure}

Figure~\ref{shapes} shows the selected shape transfers including gender, body anthropometrics and clothes changes, namely:
\begin{enumerate}
    \item female to male,
    \item male to female,
    \item corpulent to skinny,
    \item skinny to corpulent, and
    \item naked to dressed.
\end{enumerate}
 After 4 training trials, participants performed in total 45 trials. Trials were presented by block of shape transfer. At the beginning of each block, participants were shown the T-pose of the source body shape and the target-body shape. We used a latin-square design to randomize block orders. Then, for each block, 9 trials (3 methods $\times$ 3 motions) were performed by participants in a randomized order. For each participant and each trial, dependent variables were the score of similarity regarding motion and shape. Statistical analysis was performed in R using repeated-measures ART-ANOVAs and post-hoc Tukey tests. The level of significance was set to $0.05$.

\subsection{Results}

\begin{figure}
   \centering
    \includegraphics[width=0.9\columnwidth]{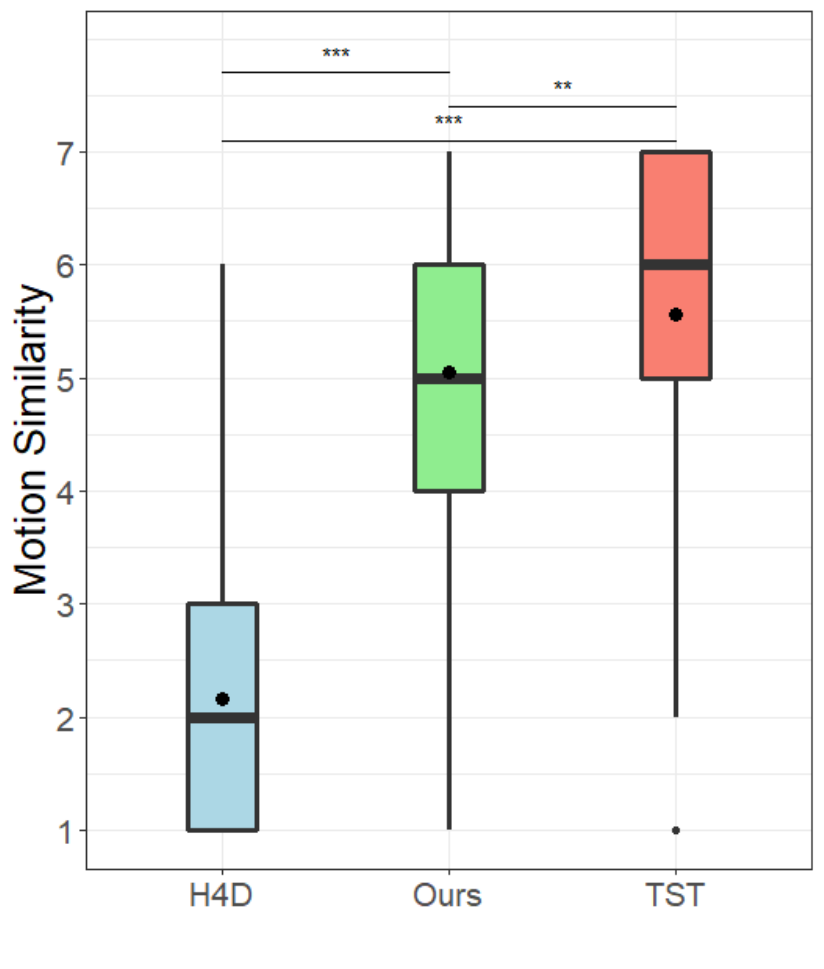}
    \includegraphics[width=0.9\columnwidth]{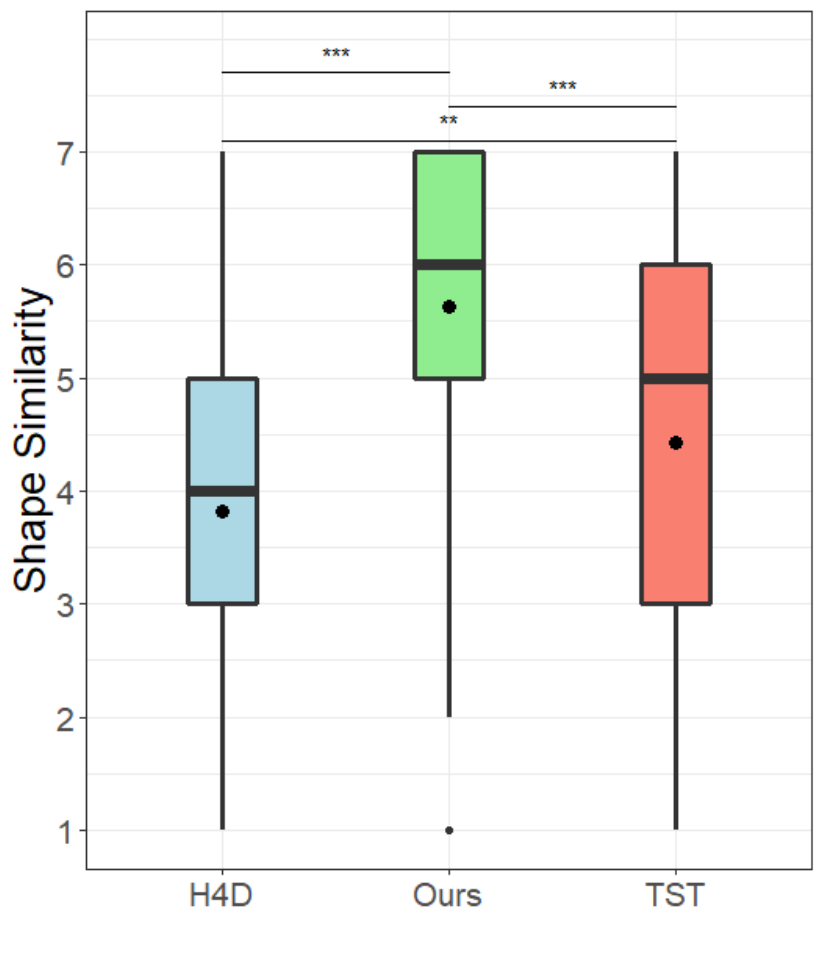}
    \caption{Results of the user study. Top: boxplots illustrating scores given by participants for motion similarity between source motion and retargeting result. Bottom: boxplots illustrating scores given by participants on shape similarity between target shape and retargeting result. Stars symbols highlight significant differences between models from the post-hoc tests ($** : p<0.001, *** : p<0.0001$).}
    
    \label{fig:motionShapeSimilarityResults}
\end{figure}

Our results show a main effect of the retargeting model both for the perceived motion ($F(2,58)=297.9, p<0.0001$) and the shape similarity ($F(2,58)=85.3, p<0.0001$). The effect is important since the effect size $\eta^{2}_{p}$ was respectively $0.91$ and $0.74$. As illustrated in Figure~\ref{fig:motionShapeSimilarityResults}, post-hoc tests show that our method was better rated than H4D both for motion ($t(58)=19.01, p<0.0001$) and shape similarity ($t(58)=12.74, p<0.0001$) in comparison to the source motion and target shape. In addition, while our method had lower scores than TST for the perceived motion ($t(58)=3.75, p<0.001$), it outperformed TST for the perceived shape ($t(58)=8.87, p<0.0001$).

\section{Potential negative societal impact}

We present a method for long-term and geometrically detailed motion retargeting between different digitized human models. It could be used without the consent of the user to animate static 3D scans, or even 3D reconstructions generated from 2D images,~\eg to generate disinformation.

%{
   %\small
  % \bibliographystyle{ieeenat_fullname}
 % \bibliography{main}
%}

\end{document}